\pdfoutput=1

\documentclass[11pt]{article}

\usepackage[preprint]{acl}

\usepackage{times}
\usepackage{latexsym}

\usepackage[T2A,T1]{fontenc}

\usepackage[utf8]{inputenc}
\usepackage[russian,english]{babel}

\usepackage{microtype}

\usepackage{inconsolata}
\usepackage{todonotes}
\usepackage{booktabs}
\usepackage{graphicx}
\usepackage{multirow}
\usepackage{adjustbox}
\title{Definition generation for lexical semantic change detection}

 \author{Mariia Fedorova, Andrey Kutuzov, Yves Scherrer \\
         University of Oslo \\ Norway \\ \texttt{\{mariiaf,andreku,yvessc\}@ifi.uio.no} }
\newcommand{\comment}[1]{}
\comment{
\author{Mariia Fedorova \\
  University of Oslo \\
  Norway \\
  \texttt{mariiaf@ifi.uio.no} \\\And
  Andrey Kutuzov \\
  University of Oslo \\
  Norway \\
  \texttt{andreku@ifi.uio.no} \\\And
  Yves Scherrer \\
  University of Oslo \\
  Norway \\
  \texttt{yves.scherrer@ifi.uio.no}
  }
 }

\begin{document}
\maketitle
\begin{abstract}
We use contextualized word definitions generated by large language models as semantic representations in the task of diachronic lexical semantic change detection (LSCD). In short, generated definitions are used as `senses', and the change score of a target word is retrieved by comparing their distributions in two time periods under comparison. On the material of five datasets and three languages, we show that generated definitions are indeed specific and  general enough to convey a signal sufficient to rank sets of words by the degree of their semantic change over time. Our approach is on par with or outperforms prior non-supervised sense-based LSCD methods. At the same time, it preserves interpretability and allows to  inspect the reasons behind a specific shift in terms of discrete definitions-as-senses. This is another step in the direction of explainable semantic change modeling.
\end{abstract}

\section{Introduction and related work}

Lexical semantic change detection (LSCD) methods up to now have mostly been based on token embeddings produced by large language models. While efficient, when measured on the existing benchmarks like diachronic word usage graphs \cite{schlechtweg-etal-2021-dwug}, these methods are largely non-interpretable and produce rather abstract `change scores'. On the other hand, historical linguistics usually deals with semantic change in terms of discrete and interpretable \textbf{senses} being lost or gained (or changing their frequency).

Recently, a number of works were published which made an attempt to bridge this gap. In particular, \citet{tang-etal-2023-word} proposed a sense distribution based LSCD method. Basically, they perform word sense disambiguation (WSD) on every occurrence of a target word in two diachronic corpora, using pre-trained sense embeddings (based on WordNet and BabelNet). Once all the occurrences are assigned a sense, the sense frequency distributions are compared between two time periods to quantify the semantic change. This approach preserves the possibility to interpret these shifts, e.g., by analyzing which sense is `responsible' for the shift.

We argue that while such methods constitute a significant advance for LSCD, they are inherently limited by their reliance on a pre-defined sense inventory. Even the best ontologies like BabelNet can miss important senses, especially when dealing with chronologically recent text data.  For many languages, good ontologies simply do not exist.

Thus, we propose to replace retrieving a fitting sense for a given target word usage from an external ontology by \textit{generating a dictionary-like contextualized definition for this specific occurrence}, using a large language model (LLM). These definitions serve as semantic representations of target word usages 
in the LSCD pipeline. The usage of generated definitions as semantic representations in LSCD was first proposed by \citet{giulianelli-etal-2023-interpretable} and further developed by \citet{kutuzov-etal-2024-enriching-word}, but they did not conduct comprehensive empirical evaluations for semantic change detection \textit{per se}. In this paper, we fill in this gap and actually test definitions as representations on the existing diachronic semantic change benchmarks. We show that our method yields competitive results, often outperforming \citet{tang-etal-2023-word}, without relying on any manually created lexical database, but at the same time preserves interpretability via human-readable definitions of senses.

The contributions of this paper are as follows:
\begin{enumerate}
    \item Contextualized definitions generated by LLMs can be used to rank words by the degree of their diachronic semantic change, with competitive performance.
    \item Using \textit{definition} embeddings with classical LSCD methods gives better results than using contextualized \textit{token} embeddings as in prior work. However, this approach makes it less convenient to interpret and analyze semantic shifts.
    \item Using generated definitions as \textit{text strings} (with some merging based on their form) yields slightly lower results in comparison, but allows inspecting the nature of a semantic shift: e.g., what senses appeared, disappeared, or changed their frequency significantly.
\end{enumerate}

All our code is available at \url{https://github.com/ltgoslo/Definition-generation-for-LSCD}.

\section{Data}

\begin{table*}
\centering
\begin{tabular}{l|ccc}
\toprule
\textbf{Strategy} & \textbf{BLEU} & \textbf{RougeL} & \textbf{BertScore} \\
\midrule
Greedy decoding & 7.384 / 6.237 / 5.113 & 0.223 / 0.198 / 0.130 & 0.860 / 0.735 / 0.700 \\
Repetition penalty (1.2) & 6.401 / 6.026 / 5.599 & 0.204 / 0.200 / 0.145 & 0.856 / 0.737 / 0.706 \\
Multinomial sampling & 6.745 / 6.037 / 4.853 & 0.200 / 0.198 / 0.122 & 0.855 / 0.736 / 0.697 \\
Beam search (5 beams) & 7.052 / \textbf{7.523} / \textbf{5.863} & 0.219 / \textbf{0.246} / \textbf{0.154} & 0.860 / 0.747 / 0.709 \\
Diverse beam search & \textbf{7.651} / 7.356 / 5.713 & \textbf{0.225}  / 0.243 / 0.150 & \textbf{0.862} / \textbf{0.750} / \textbf{0.710} \\
\bottomrule
\end{tabular}
\caption{Performance of English / Norwegian / Russian definition generation with different generation strategies.}
\label{tab:defgen}
\end{table*}

We experiment on English, Norwegian and Russian benchmarks, since for these languages, definition generators were already created by \citet{kutuzov-etal-2024-enriching-word}. However, scaling to other languages is comparatively easy and requires only a small dataset of contextualized definitions (see §\ref{sec:generation}).

To evaluate the performance of a semantic change detection system, we used existing LSCD datasets (diachronic corpora and gold scores for the target words): the English part of the Sem\-Eval'20 Task 1\footnote{\url{https://www.ims.uni-stuttgart.de/en/research/resources/corpora/sem-eval-ulscd/}} \cite{schlechtweg-etal-2020-semeval}, NorDiaChange\footnote{\url{https://github.com/ltgoslo/nor_dia_change}} \cite{kutuzov-etal-2022-nordiachange} for Norwegian, and RuShiftEval\footnote{\url{https://github.com/akutuzov/rushifteval_public/tree/main}} \cite{kutuzov-pivovarova-2021-three} for Russian. NorDiaChange actually contains two datasets and RuShiftEval contains three datasets, with different time period pairs under comparison (for Norwegian, the sets of target words are also different). The Russian datasets feature the highest number of target words (99, as compared to 37 in English and Norwegian datasets).

Note that SemEval'20 Task 1 included two subtasks: binary classification of words (changed or not changed) and ranking the words by the degree of their change. In this work, we focus only on the ranking task: 1) because the Russian dataset does not include binary labels, and 2) because even in the English and Norwegian datasets the binary labels are in many ways derivatives of the change scores.\footnote{In contrast to the English and Norwegian datasets which contain \textit{change} scores, the Russian datasets contain \textit{similarity} scores. The obtained correlations are thus negative. We flip the sign when reporting these numbers to improve readability.}

It is also important to note that the Ru\-Shift\-Eval dataset was used in a shared task of the same name \cite{rushifteval2021}. However, the scores in its leaderboard or in \citet{cassotti-etal-2023-xl} are not directly comparable to the scores in this work, since in the shared task, the dataset was split into development and test parts, so that the participants were able to tune their systems on the development set. In this paper, we focus on unsupervised approaches, aiming to avoid the necessity of tuning hyperparameters and leaving this for future work.

\subsection{Preprocessing}

We use the lemmatized versions of the SemEval-2020 English corpora when reproducing \citet{tang-etal-2023-word}'s Lesk baseline. 
No preprocessing of the Norwegian and Russian corpora has been done, except for lower-casing when running the Lesk baselines (see the details in the section~\ref{sec:experiments}) and taking lemmas of the target words into account when sampling usage examples for both Lesk and definition generation methods. Since frequent words may have more than 100 000 occurrences in the Norwegian and Russian corpora, we sampled randomly no more than 1000 usages (sentences) for each target word from every diachronic corpus. 

This resulted in sub-corpora of total 58\,000 word usages for English, 47\,000 for Norwegian-1, 51\,000 for Norwegian-2, and 164\,000, 183\,000 and 168\,000 for Russian-1, Russian-2 and Russian-3 correspondingly.\footnote{We use only examples no longer than 350 subword tokens in all our experiments.}

\section{Definition generation methods}
\label{sec:generation}

Our general pipeline of generating definitions  from an LLM (`DefGen') is similar to \citet{giulianelli-etal-2023-interpretable}. We use the definition generation models presented by \citet{kutuzov-etal-2024-enriching-word}. They are based on \texttt{mT0-xl}~\cite{muennighoff-etal-2023-crosslingual} and were fine-tuned on \textbf{WordNet}~\cite{ishiwatari-etal-2019-learning}, \textbf{Oxford}~\cite{gadetsky-etal-2018-conditional} and \textbf{CoDWoE}~\cite{mickus-etal-2022-semeval} for English, \textbf{CoDWoE} for Russian and  \textbf{Bokmålsordboka}\footnote{\url{https://ordbokene.no/}} for Norwegian.  The models are extensively described and evaluated in \citet{kutuzov-etal-2024-enriching-word}, but we provide the most important details in Appendix~\ref{sec:model_desc}.

All CoDWoE datasets originally come from Wiktionary, so it is straightforward to extend this method to any major language. As a prompt for the LLM, we used the original example usage followed by the question `What is the definition of TARGETWORD?' (in English, Norwegian or Russian).

Additionally, we  conducted a series of experiments with different generation strategies. Prior work used only basic greedy decoding, while we experimented with alternative strategies such as multinomial sampling, beam search, and diverse beam search \cite{divbeam}.

Note that these experiments do not deal with LSCD -- they only evaluate the capability of the models to generate definitions similar to the gold ones. The results of our experiments for English, Russian and Norwegian, evaluated with BLEU, RougeL and BertScore, are shown in Table~\ref{tab:defgen}. We used the default implementations of these metrics from the \texttt{Evaluate} library\footnote{\url{https://huggingface.co/docs/evaluate/}} with the only change of using whitespace tokenizer in RougeL for all languages (instead of the default one aimed at English). For English and Russian, we evaluated on the CoDWoE trial sets (about 200 instances each); for Norwegian, we used our own test set of about 7000 instances.\footnote{Generating definitions with our models for 1000 example usages takes about 2 minutes on an NVIDIA A100 GPU.}

The performance scores are consistent across all three languages: the default mode of greedy decoding turned to be a hard-to-beat baseline. However, using beam search with 5 beams  (or its diverse version with diversity penalty of 0.5) does outperform greedy decoding according to all three metrics.

In the experiments below, we use definitions generated with all three approaches: greedy decoding, beam search and diverse beam search, to explore to what extent the definition generation performance translates to LSCD performance.

\section{Semantic change detection with generated definitions}
\label{sec:experiments}

In this section, we describe the methods we used in our experiments, starting with the XLM-R and sense embeddings baselines from prior work. Next, we present our own approach based on definition embeddings and the standard LSCD aggregation algorithms: PRT (prototype embeddings), APD (average pairwise distance), and their arithmetic mean (AM) \cite{kutuzov-etal-2022-contextualized}.  Finally, we move on to our \textit{interpretable} `definitions-as-senses` method which employs definitions directly (as text strings) and merges similar definitions together using Levenshtein edit distance.

\begin{table*}
\adjustbox{max width=\textwidth}{%
\begin{tabular}{lllllll}
\toprule
\textbf{Method} & \textbf{English} & \textbf{Norwegian-1} & \textbf{Norwegian-2} & \textbf{Russian-1} & \textbf{Russian-2} & \textbf{Russian-3}\\
\midrule
\multicolumn{7}{c}{\textbf{Non-interpretable methods:}} \\
\midrule
XLM-R token embeddings & 0.514$^\diamondsuit$ & 0.394$^\diamondsuit$ & 0.387$^\diamondsuit$ & 0.376$^\diamondsuit$ & \textbf{0.480}$^\diamondsuit$ & 0.457$^\diamondsuit$ \\
\midrule
Definition embeddings (ours) & \textbf{0.637} & \textbf{0.496} & \textbf{0.565} & \textbf{0.488} & 0.462 & \textbf{0.504} \\
(See §~\ref{subsec:defemb} / Table~\ref{tab:apd-prt} for details) \\
\midrule
\midrule
\multicolumn{7}{c}{\textbf{Interpretable methods:}} \\
\midrule
Lesk without PoS & 0.423$^\clubsuit$ & \textit{0.178} & 0.500 & \multirow{2}{*}{0.294} & \multirow{2}{*}{0.279} & \multirow{2}{*}{0.286} \\
Lesk with PoS & 0.587 & \textit{0.150} & 0.474 &  &  &  \\
\midrule
ARES sense embeddings & 0.529$^\clubsuit$ & --- & --- & --- & --- & --- \\
LMMS sense embeddings & 0.589$^\clubsuit$ & --- & --- & --- & --- & --- \\
\midrule
Definitions as senses (ours) & 0.565 & 0.362 & \textit{0.260} & 0.391 & 0.431 & 0.491 \\
(See §~\ref{subsec:defmerge} / Table~\ref{tab:merge} for details) & \\
\bottomrule
\end{tabular}}
\caption{Summary of our results and baselines (Spearman's $\rho$ for graded LSCD). Figures marked with $\diamondsuit$ are best results from \citet{giulianelli-etal-2022-fire}. Figures marked with $\clubsuit$ are taken from \citet{tang-etal-2023-word}; Lesk is called NLTK in their paper. Numbers without a symbol are our own results. All correlations are statistically significant with $p < 0.05$, except Norwegian-2 with definitions as senses and Norwegian-1 with Lesk (those are marked in italics).}
\label{tab:baselines}
\end{table*}

\subsection{Baselines}

Table~\ref{tab:baselines} shows the results from previous studies that we use as our baselines, as well as the best results of our definition-based systems.

The first baseline is an LSCD approach based on contextualized \textbf{XLM-R token embeddings} \cite{giulianelli-etal-2022-fire} (best scores).  It is not interpretable, but has reliably high performance across all languages,  according to \citet{periti2024systematic}; we refer the reader to this survey for additional LSCD baselines with contextualized word embeddings.

As a second, interpretable baseline, we follow \citet{tang-etal-2023-word} and start with the \textbf{Lesk} WSD algorithm \citep{lesk1986automatic} with WordNet definitions (this method is called `NLTK' in the Table 1 of their paper). Their result for English, as well as our extensions to Norwegian and Russian, are shown in the lower part of Table~\ref{tab:baselines}. We were able to reproduce their Lesk results with only small fluctuations\footnote{Probably due to the fact that we used top 1 sense in all our experiments, while \citet{tang-etal-2023-word} experimented with top k highest ranked senses on a held-out set and found k = 2 to perform best. However, we focus on unsupervised approaches to the task and leave hyperparameter tuning on development sets for future work.}. Since no open WordNet-like databases exist for Norwegian or Russian\footnote{The Open Multilingual WordNet allows searching for words in other languages than English, but the synset definitions remain in English.}, we used the aforementioned Bokmålsordboka and CoDWoE/Wiktionary as sources of Norwegian and Russian sense definitions.

We also experimented with adding part-of-speech information to the Lesk algorithm (that is, restricting Lesk WSD search to only the synsets corresponding to the desired part-of-speech of the target word). The English SemEval'20 dataset explicitly specifies parts-of-speech for the target words, while the Norwegian and Russian datasets contain nouns only. The two variants of Lesk yield identical results on the Russian datasets since the target words are not PoS-ambiguous. For English and Norwegian-2, Lesk outperforms the XLM-R token embeddings and comes close to our approach based on definition embeddings. 

However, \citet{tang-etal-2023-word}'s main results are based on \textbf{ARES and LMMS sense embeddings}. Unfortunately, these embeddings are not publicly available anymore due to link rot. The LMMS download link\footnote{\url{https://github.com/danlou/LMMS}} leads to a private file storage, and the ARES embeddings also cannot be found at the provided URL.
We contacted the authors but got no answer by the time of writing. Thus, we can only quote the performance scores from \citet{tang-etal-2023-word}. 

\subsection{Using definition embeddings}
\label{subsec:defemb}

\begin{table*}
\adjustbox{max width=\textwidth}{%
\begin{tabular}{llllllllll}
\toprule
 & \multicolumn{3}{c}{\textbf{English}} & \multicolumn{3}{c}{\textbf{Norwegian-1}} &  \multicolumn{3}{c}{\textbf{Norwegian-2}} \\
\cmidrule(lr){2-4} \cmidrule(lr){5-7} \cmidrule(lr){8-10}
 & APD & PRT & AM & APD & PRT & AM & APD & PRT & AM \\
\midrule
\textbf{Token emb.} & 0.514$^\diamondsuit$ & 0.320$^\diamondsuit$ & 0.457$^\diamondsuit$ & 0.389$^\diamondsuit$ & 0.378$^\diamondsuit$ & 0.394$^\diamondsuit$ & 0.387$^\diamondsuit$ & 0.270$^\diamondsuit$ & 0.325$^\diamondsuit$ \\
\textbf{Greedy} (ours) & 0.633 & 0.331 & 0.580 & 0.416 & 0.368 & \bf 0.496 & \bf 0.565 & 0.413 & 0.558 \\
\textbf{Beam} (ours) & \bf 0.637 & 0.355 & 0.601 & \it 0.317 & 0.411 & 0.434 & 0.478 & 0.452 & 0.479 \\
\textbf{Diverse} (ours) & 0.613 & 0.359 & 0.591 & 0.335 & 0.364 & 0.444 & 0.508 & 0.470 & 0.523 \\
\midrule
& \multicolumn{3}{c}{\textbf{Russian-1}} & \multicolumn{3}{c}{\textbf{Russian-2}} & \multicolumn{3}{c}{\textbf{Russian-3}} \\
\cmidrule(lr){2-4} \cmidrule(lr){5-7} \cmidrule(lr){8-10}
 & APD & PRT & AM & APD & PRT & AM & APD & PRT & AM \\
\midrule
\textbf{Token emb.} & 0.372$^\diamondsuit$ & 0.294$^\diamondsuit$ & 0.376$^\diamondsuit$ & \bf 0.480$^\diamondsuit$ & 0.313$^\diamondsuit$ & 0.374$^\diamondsuit$ & 0.457$^\diamondsuit$ & 0.313$^\diamondsuit$ & 0.384$^\diamondsuit$ \\
\textbf{Greedy} (ours) & 0.464 & 0.406 & \bf 0.488 & 0.453 & 0.430 & 0.462 & 0.489 & \bf 0.504 & 0.494 \\
\textbf{Beam} (ours) & 0.381 & 0.387 & 0.401 & 0.400 & 0.451 & 0.411 & 0.386 & 0.439 & 0.413 \\
\textbf{Diverse} (ours) & 0.396 & 0.457 & 0.433 & 0.405 & 0.449 & 0.417 & 0.414 & 0.476 & 0.436 \\
\bottomrule
\end{tabular}}
\caption{LSCD performance (Spearman's $\rho$) with definition embeddings obtained with different decoding strategies (greedy decoding, beam search and diverse beam search). For comparison, \textit{Token emb.} presents the results by \citet{giulianelli-etal-2022-fire} with contextualized XLM-R token embeddings. AM is the arithmetic mean of APD and PRT. All correlations except one (Norwegian-1 APD with beam search, marked in italics) are statistically significant with $p < 0.05$.}
\label{tab:apd-prt}
\end{table*}

Generated definitions can be easily vectorized by using any sentence embedding model. We embedded the generated definitions for every target word usage with DistilRoBERTa\footnote{\url{https://huggingface.co/sentence-transformers/all-distilroberta-v1}}. After that, it becomes possible to use the standard LSCD methods like PRT,  APD, and their arithmetic mean (AM) \cite{kutuzov-etal-2022-contextualized}. The only difference to the standard setup is that instead of \textit{token} embeddings, we feed contextualized \textit{definition} embeddings into the algorithm.

The intuition here is that by measuring the average or pairwise distances between definitions of one and the same target word in two historical corpora, one can quantify the degree of semantic change for this word between two time periods. As can be seen in the Table~\ref{tab:apd-prt}, this is indeed the case. Our definition embeddings outperform the contextualized XLM-R token embeddings from \citet{giulianelli-etal-2022-fire} on five of the six evaluated datasets.

Note in this context that using token embeddings from a masked LM requires the knowledge of the exact position of the target token in the input sentence (with additional issues in case of the target word being split into multiple sub-words). In our approach, adding the `What is the definition of X?' prompt to the input sentence is completely decoupled from the location of X within the sentence.

The decoding strategy does not seem to make a significant difference in terms of LSCD performance. Greedy decoding is a reasonable default choice despite its slightly lower scores in Table~\ref{tab:defgen}.

On English, the APD method on definition embeddings also outperforms the best sense-embedding-based approaches from \citet{tang-etal-2023-word} by a large margin (see Table~\ref{tab:baselines}). Note, however, that using definition embeddings in this case still yields a non-interpretable result: we do not know what exact senses are responsible for a high degree of semantic change. For this reason, we propose to use the generated definitions directly in the next section.

\subsection{Merging definitions together}
\label{subsec:defmerge}

The definitions generated by a DefGen system can be used directly for LSCD. In this case, each unique definition is considered a separate word sense, and the sense distributions of the two time periods can be compared in the same way as in \citet{tang-etal-2023-word}. This approach is straightforward and already results in competitive performance (see the ``No merging'' section in Table~\ref{tab:merge}\footnote{Table~\ref{tab:merge} reports results after using two different distance metrics: cosine and Jensen-Shannon divergence (JSD). JSD is superior in most cases, but not always.}).

However, it obviously suffers from too granular senses. As an example, for almost 1000 occurrences of the word `\textit{plane}' in the SemEval'20 English dataset, more than 200 unique definitions were generated, most with only one occurrence. This list includes definitions obviously belonging to one and the same sense: for example, `\textit{An aircraft, especially one designed for military use}' and  `\textit{An aircraft, especially a military aircraft}'. This leads to noise and -- even worse -- to reduced interpretability. It is easy to observe that definitions belonging to the same sense are often similar in their surface form. Thus, in this subsection, we describe our experiments with merging similar definitions together.

\begin{table*}
\adjustbox{max width=\textwidth}{%
\begin{tabular}{lrr|rrrr|rrrrrr}
\toprule
& \multicolumn{2}{c|}{\textbf{English}} & \multicolumn{2}{c}{\textbf{Norwegian-1}} & \multicolumn{2}{c}{\textbf{Norwegian-2}} & \multicolumn{2}{|c}{\textbf{Russian-1}} & \multicolumn{2}{c}{\textbf{Russian-2}} & \multicolumn{2}{c}{\textbf{Russian-3}} \\
\cmidrule(lr){2-3} \cmidrule(lr){4-5} \cmidrule(lr){6-7} \cmidrule(lr){8-9} \cmidrule(lr){10-11} \cmidrule(lr){12-13}
& Cosine & JS & Cosine & JS & Cosine & JS & Cosine & JS & Cosine & JS & Cosine & JS \\
\midrule
\multicolumn{13}{c}{\bf No merging:} \\
\textbf{Greedy} & 0.461 & 0.405 & \textit{0.303} & 0.332 & \textit{0.211} & \textit{0.232} & 0.299 & 0.390 & 0.337 & 0.427 & 0.383 & 0.469 \\
\textbf{Beam} & 0.457 & 0.476 & \textit{0.268} & \textit{0.238} & \textit{0.216} & \textit{0.201} & 0.304 & 0.368 & 0.297 & 0.403 & 0.317 & 0.417 \\
\textbf{Diverse} & 0.449 & 0.382 & \textit{0.241} & \textit{0.280} & \textit{0.069} & \textit{0.164} & 0.301 & 0.345 & 0.310 & 0.389 & 0.348 & 0.421\\
\midrule
\multicolumn{13}{c}{\bf Minimalist merging:} \\
\textbf{Greedy} & 0.564 & \bf 0.565 & \textit{0.251} & \textit{0.280} & \textit{0.192} & \textit{0.197} & 0.271 & \textbf{0.391} & 0.233 & \textbf{0.431} & 0.325 & \textbf{0.491}  \\
\textbf{Beam} & 0.510 & 0.500 & \textit{0.297} & \textit{0.240} & \textit{0.112} & \textit{0.189} & 0.298 & 0.366 & 0.252 & 0.383 & 0.301 & 0.409\\
\textbf{Diverse} & 0.478 & 0.434 & 0.325 & \textit{0.296} & \textit{0.162} & \textit{0.215} & 0.265 & 0.354 & 0.268 & 0.406 & 0.287  & 0.443\\
\midrule
\multicolumn{13}{c}{\bf Full-fledged merging:} \\
\textbf{Greedy} & 0.417 & 0.418 & \textit{0.261} & \bf 0.362 & \textit{0.193} & \bf \textit{0.260} & 0.286 & \textbf{0.391} & 0.250 &  0.416 & 0.360 & 0.476  \\
\textbf{Beam} & 0.492 & 0.493 & \textit{0.265} & \textit{0.215} & \textit{0.186} & \textit{0.226} & 0.304 & 0.360 & 0.250 & 0.347 & 0.327  & 0.420 \\
\textbf{Diverse} & \textit{0.312} & \textit{0.301} & \textit{0.209} & \textit{0.315} & \textit{0.202} & \textit{0.221} & 0.236 & 0.301 & 0.217 & 0.379 & 0.262 & 0.411  \\
\midrule
\textbf{Threshold} & \multicolumn{2}{c}{50} & \multicolumn{2}{c}{10} & \multicolumn{2}{c}{10} & \multicolumn{2}{c}{10} & \multicolumn{2}{c}{10} & \multicolumn{2}{c}{10} \\
\bottomrule
\end{tabular}}
\caption{LSCD performance (Spearman's $\rho$) with generated definitions and different generation and merging strategies. Results are reported with two distance metrics: cosine similarity and Jensen-Shannon divergence. \textit{Threshold} refers to the Levenshtein edit distance threshold used for merging definitions. All correlations are statistically significant with $p < 0.05$, except those marked in italics.}
\label{tab:merge}
\end{table*}

Any decision about what word usages belong to one sense is inherently arbitrary \cite{kilgarriff1997don}. The same applies to definitions: in order to decide whether two definitions represent one and the same sense, one has to find a way to quantify their similarity. In order to preserve interpretability, we decided to use surface string similarity metrics (as opposed to, e.g., cosine similarity between definition embeddings).

We remind again that the top part of Table~\ref{tab:merge} shows the performance scores on our datasets with no merging involved: every unique definition is considered to be a separate sense on its own and we simply compare the distribution of these `senses' across two time periods. In addition to that, we introduce two merging strategies which we dub `minimalist' and `full-fledged' merging. The intuition behind them is that one replaces some of the generated definitions for a target word with another \textit{similar} definition generated for the same target word, thus reducing the total number of unique definitions per word and making it closer to a realistic number of senses.

First, we filter out punctuation marks from all definitions. Second, every time period (out of two) is processed \textit{separately}. We also tried \textit{joint} processing of both time periods to make the resulting definitions-as-senses more comparable. However, it consistently resulted in worse performance: the most probable reason being that it makes the sense distributions too close to each other, eliminating meaningful differences. It also can bias the predictions if one time period is represented by a larger corpus than another.

\begin{figure}
\centering
\includegraphics[width=\linewidth]{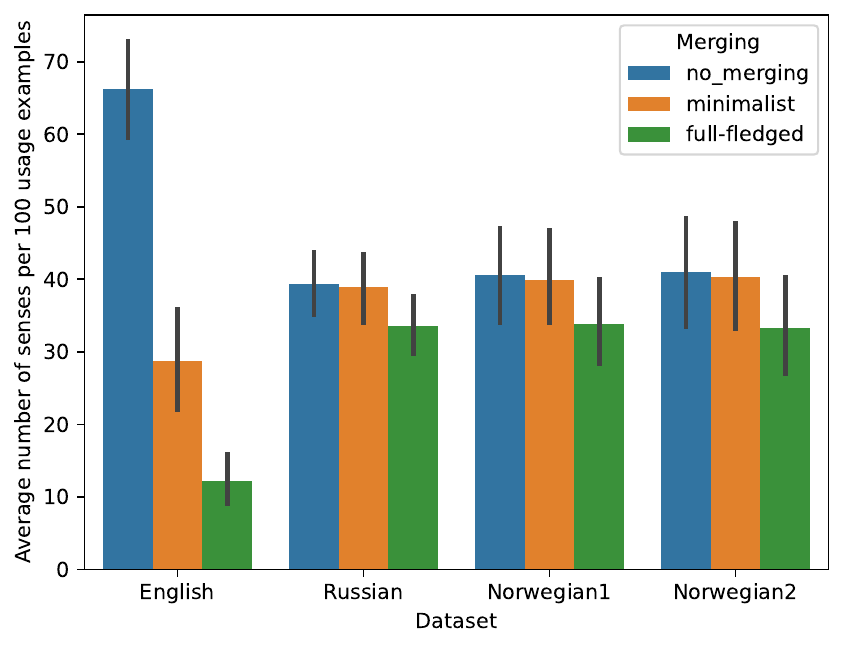}
\caption{Average number of senses per 100 usages before and after merging, calculated across all datasets for each language.}
\label{fig:merged_number_of_senses}
\end{figure}

The processing is done as follows. For every target word, we sort the generated definitions by their frequency and loop over them, starting from the top (most frequent) ones, representing the dominant senses of the word. For every step in this loop (let's designate it as `hub definition'), we loop again over the remaining definitions, calculating the edit distance\footnote{We use Levenshtein distance \cite{levenshtein}; other edit distance formulations are possible. In addition, only definitions containing 4 words and more could serve as hubs, to prevent merging together very short definitions with low edit distances.} between them and the hub definition. If the edit distance is lower than the pre-defined threshold, the current definition is replaced with the hub definition (we assume they belong to one sense). With the `minimalist strategy', only the first (most frequent) definition can be the hub (and have other definitions replaced by it), the loop is stopped after it is compared to all other definitions. With the `full-fledged' strategy, the loop continues, and other (less frequent) definitions also get a chance to become hubs, if they were not subsumed by another definition before. The `full-fledged' strategy naturally results in even stronger reduction on the number of unique senses (see Figure~\ref{fig:merged_number_of_senses}). The datasets with replaced definitions are used for LSCD in the usual way.

The value of the edit distance threshold is a hyperparameter. One can tune it on a designated development set, but in this study, we tried to avoid the supervised setup. Thus, after studying the data, we only tested two intuitively sensible threshold values of $10$ and $50$. It turned out that the value of $10$ is optimal for Norwegian and Russian, while the value of $50$ (more merging) is optimal for English (changes in the number of senses after merging are more obvious for  English on Figure~\ref{fig:merged_number_of_senses} because of the more aggressive Levenshtein threshold).

As shown in Table~\ref{tab:merge}, merging definitions indeed improves the performance in the LSCD ranking task for all the languages and benchmarking datasets, in comparison to no merging at all. It still does not reach the level of APD on definition embeddings (Table~\ref{tab:apd-prt}), but outperforms one of two sense embedding approaches from \citet{tang-etal-2023-word}. For two Russian datasets out of three, our merged definitions-as-senses outperform previous best unsupervised results \cite{giulianelli-etal-2022-fire}.

Curiously, only for Norwegian the best performance is achieved using the full-fledged merging strategy (and is still comparatively low); for English and Russian, the minimalist strategy (only merging with the dominant sense) gives the best scores. Thus, merging should be done cautiously: merging too much can degrade the performance. Another interesting finding is that greedy decoding again turned out to be the best generation strategy for LSCD. We hypothesize that using beam search often results in too diverse a set of definitions, which prevents efficient comparisons between their diachronic distributions.

Overall, using surface forms of generated definitions directly is outperformed by using vectorized definitions and APD or PRT methods for semantic change detection (see \ref{subsec:defemb}), which is consistent with findings in \citet{giulianelli-etal-2023-interpretable}. They found that cosine similarity between vectorized definitions better approximates human similarity judgments than surface form similarity metrics like edit distance or BLEU. However, using definitions in their textual form has a clear advantage of being interpretable. In the next section, we illustrate how semantic change can be explained and analyzed by comparing the generated (and merged) definitions.

\section{Qualitative analysis}
\label{sec:analysis}

When predicting semantic change on the basis of the distribution of senses (or definitions-as-senses), it becomes possible to  analyze and interpret this change, by simply looking at the distribution of entries (senses) which contribute most to the difference.

Let's consider the top performing set of English definitions (generated with greedy decoding and merged in the minimalist approach with the edit distance threshold of $50$). For the word `\textbf{ball}' in the SemEval'20 time periods, the JSD metric yields a high change score of $0.83$. After looking at the list of top frequent definitions-as-senses for this word, it becomes clear that its dominant sense has changed: while in time period 1 (19th century), more than 82\% of all usage examples were given the definition \textit{`A spherical object especially one that is round in shape'} with \textit{`A party'} being the next most frequent sense, in time period 2 (20th century), 80\% of `\textbf{ball}' usages were defined as \textit{`The object hit in a game'}, with similar definitions following this one in the top-frequent list. This is a clear evidence of the `dancing party' sense for the word `ball' becoming obsolete in the 20th century, with the sports-related sense taking the dominant position.

For the noun `\textbf{attack}', the system predicts a medium change degree of $0.34$. Again, it is straightforward to find the reasons. While in both periods the dominant sense is the same (\textit{`An instance of military action against an enemy'}), in the time period 2 its ratio drops down from 87\% to 80\%, and we observe the appearance of a new rare but not unique sense of \textit{`An instance of sudden violent activity of a bodily organ or system especially the heart'}. This is a linguistically plausible explanation of a semantic shift, much more useful to a lexicographer than a raw change score. See Appendix~\ref{sec:defex} for more examples in English, Norwegian and Russian.

\section{Conclusion}

In this paper, we showed how contextualized dictionary-like definitions generated by a fine-tuned large language model can be used for the practical downstream task of semantic change detection (in particular, ranking words by the degree of their diachronic semantic change).

Following \citet{giulianelli-etal-2023-interpretable}, we treat generated definitions as semantic representations of the target words. These definitions (and their frequency distributions) can be used `as is', following \citet{tang-etal-2023-word}'s method, or after embedding them in a dense vector space using any available sentence embedding model. The second method yields results which are empirically better (considering existing benchmarks for three different languages), but the first method makes it much easier to interpret and explain semantic change, since it operates directly on generated definitions in their textual forms.

We consider this study a small step towards more explainable semantic change modeling, which can be closer to linguistically plausible discrete `senses', while still retaining empirical performance. In the future, we plan to explore to what extent it is possible to improve our results by tuning hyperparameters on development sets (where available). Another direction for future research is using more advanced string distance metrics like weighted Levenshtein distance, Longest Common Subsequence Ratio, or Word Mover's Distance \cite{kusner2015word}, in the hope that it will allow to handle more nuanced similarities and dissimilarities between generated definitions.

\section*{Ethical impact}
For fine-tuning our definition generators, only open and publicly available datasets, mostly dictionaries were used. However, some of them (especially Wiktionary) are crowd-sourced, and thus can (and do) contain inappropriate phrases. In addition, the foundational \texttt{mT0} language model on which we base our pipeline, was trained among other data on web-crawled texts, also far from being clean. Thus, generated definitions are not guaranteed to be free from swearing, discriminative passages and other inappropriate content.

\section*{Limitations}

This work is limited to only three languages (English, Norwegian and Russian), while the standard SemEval'20 `LSCD suite' contains four languages (English, German, Latin, Swedish). Also, we did not experiment with hyperparameter tuning or different ways of training definition generators. It should also be noted that Spearman rank correlation may not be accurate for samples the size of LSCD benchmarks: we use it to preserve compatibility and comparability with prior work.

Finally, we have not yet empirically evaluated how useful in practice the definition-based explanations of semantic change will be for historical linguists and lexicographers (although what we see after manual inspection of the system predictions is promising).

\bibliography{anthology,custom}

\begin{thebibliography}{22}
\expandafter\ifx\csname natexlab\endcsname\relax\def\natexlab#1{#1}\fi

\bibitem[{Cassotti et~al.(2023)Cassotti, Siciliani, DeGemmis, Semeraro, and
  Basile}]{cassotti-etal-2023-xl}
Pierluigi Cassotti, Lucia Siciliani, Marco DeGemmis, Giovanni Semeraro, and
  Pierpaolo Basile. 2023.
\newblock \href {https://doi.org/10.18653/v1/2023.acl-short.135}
  {{XL}-{LEXEME}: {W}i{C} pretrained model for cross-lingual {LEX}ical
  s{EM}antic chang{E}}.
\newblock In \emph{Proceedings of the 61st Annual Meeting of the Association
  for Computational Linguistics (Volume 2: Short Papers)}, pages 1577--1585,
  Toronto, Canada. Association for Computational Linguistics.

\bibitem[{Chung et~al.(2022)Chung, Hou, Longpre, Zoph, Tay, Fedus, Li, Wang,
  Dehghani, Brahma et~al.}]{chung2022scaling}
Hyung~Won Chung, Le~Hou, Shayne Longpre, Barret Zoph, Yi~Tay, William Fedus,
  Eric Li, Xuezhi Wang, Mostafa Dehghani, Siddhartha Brahma, et~al. 2022.
\newblock Scaling instruction-finetuned language models.
\newblock \emph{arXiv preprint arXiv:2210.11416}.

\bibitem[{Gadetsky et~al.(2018)Gadetsky, Yakubovskiy, and
  Vetrov}]{gadetsky-etal-2018-conditional}
Artyom Gadetsky, Ilya Yakubovskiy, and Dmitry Vetrov. 2018.
\newblock \href {https://doi.org/10.18653/v1/P18-2043} {Conditional generators
  of words definitions}.
\newblock In \emph{Proceedings of the 56th Annual Meeting of the Association
  for Computational Linguistics (Volume 2: Short Papers)}, pages 266--271,
  Melbourne, Australia. Association for Computational Linguistics.

\bibitem[{Giulianelli et~al.(2022)Giulianelli, Kutuzov, and
  Pivovarova}]{giulianelli-etal-2022-fire}
Mario Giulianelli, Andrey Kutuzov, and Lidia Pivovarova. 2022.
\newblock \href {https://doi.org/10.18653/v1/2022.lchange-1.6} {Do not fire the
  linguist: Grammatical profiles help language models detect semantic change}.
\newblock In \emph{Proceedings of the 3rd Workshop on Computational Approaches
  to Historical Language Change}, pages 54--67, Dublin, Ireland. Association
  for Computational Linguistics.

\bibitem[{Giulianelli et~al.(2023)Giulianelli, Luden, Fernandez, and
  Kutuzov}]{giulianelli-etal-2023-interpretable}
Mario Giulianelli, Iris Luden, Raquel Fernandez, and Andrey Kutuzov. 2023.
\newblock \href {https://doi.org/10.18653/v1/2023.acl-long.176} {Interpretable
  word sense representations via definition generation: The case of semantic
  change analysis}.
\newblock In \emph{Proceedings of the 61st Annual Meeting of the Association
  for Computational Linguistics (Volume 1: Long Papers)}, pages 3130--3148,
  Toronto, Canada. Association for Computational Linguistics.

\bibitem[{Ishiwatari et~al.(2019)Ishiwatari, Hayashi, Yoshinaga, Neubig, Sato,
  Toyoda, and Kitsuregawa}]{ishiwatari-etal-2019-learning}
Shonosuke Ishiwatari, Hiroaki Hayashi, Naoki Yoshinaga, Graham Neubig, Shoetsu
  Sato, Masashi Toyoda, and Masaru Kitsuregawa. 2019.
\newblock \href {https://doi.org/10.18653/v1/N19-1350} {Learning to describe
  unknown phrases with local and global contexts}.
\newblock In \emph{Proceedings of the 2019 Conference of the North {A}merican
  Chapter of the Association for Computational Linguistics: Human Language
  Technologies, Volume 1 (Long and Short Papers)}, pages 3467--3476,
  Minneapolis, Minnesota. Association for Computational Linguistics.

\bibitem[{Kilgarriff(1997)}]{kilgarriff1997don}
Adam Kilgarriff. 1997.
\newblock I don’t believe in word senses.
\newblock \emph{Computers and the Humanities}, 31(2):91--113.

\bibitem[{Kusner et~al.(2015)Kusner, Sun, Kolkin, and
  Weinberger}]{kusner2015word}
Matt Kusner, Yu~Sun, Nicholas Kolkin, and Kilian Weinberger. 2015.
\newblock From word embeddings to document distances.
\newblock In \emph{International conference on machine learning}, pages
  957--966. PMLR.

\bibitem[{Kutuzov et~al.(2024)Kutuzov, Fedorova, Schlechtweg, and
  Arefyev}]{kutuzov-etal-2024-enriching-word}
Andrey Kutuzov, Mariia Fedorova, Dominik Schlechtweg, and Nikolay Arefyev.
  2024.
\newblock \href {https://aclanthology.org/2024.lrec-main.546} {Enriching word
  usage graphs with cluster definitions}.
\newblock In \emph{Proceedings of the 2024 Joint International Conference on
  Computational Linguistics, Language Resources and Evaluation (LREC-COLING
  2024)}, pages 6189--6198, Torino, Italia. ELRA and ICCL.

\bibitem[{Kutuzov and Pivovarova(2021{\natexlab{a}})}]{rushifteval2021}
Andrey Kutuzov and Lidia Pivovarova. 2021{\natexlab{a}}.
\newblock Ru{S}hift{E}val: a shared task on semantic shift detection for
  {R}ussian.
\newblock In \emph{Computational linguistics and intellectual technologies:
  Papers from the annual conference Dialogue}.

\bibitem[{Kutuzov and
  Pivovarova(2021{\natexlab{b}})}]{kutuzov-pivovarova-2021-three}
Andrey Kutuzov and Lidia Pivovarova. 2021{\natexlab{b}}.
\newblock \href {https://doi.org/10.18653/v1/2021.lchange-1.2} {Three-part
  diachronic semantic change dataset for {R}ussian}.
\newblock In \emph{Proceedings of the 2nd International Workshop on
  Computational Approaches to Historical Language Change 2021}, pages 7--13,
  Online. Association for Computational Linguistics.

\bibitem[{Kutuzov et~al.(2022{\natexlab{a}})Kutuzov, Touileb, M{\ae}hlum,
  Enstad, and Wittemann}]{kutuzov-etal-2022-nordiachange}
Andrey Kutuzov, Samia Touileb, Petter M{\ae}hlum, Tita Enstad, and Alexandra
  Wittemann. 2022{\natexlab{a}}.
\newblock \href {https://aclanthology.org/2022.lrec-1.274} {{N}or{D}ia{C}hange:
  Diachronic semantic change dataset for {N}orwegian}.
\newblock In \emph{Proceedings of the Thirteenth Language Resources and
  Evaluation Conference}, pages 2563--2572, Marseille, France. European
  Language Resources Association.

\bibitem[{Kutuzov et~al.(2022{\natexlab{b}})Kutuzov, Velldal, and
  {\O}vrelid}]{kutuzov-etal-2022-contextualized}
Andrey Kutuzov, Erik Velldal, and Lilja {\O}vrelid. 2022{\natexlab{b}}.
\newblock \href
  {https://doi.org/https://doi.org/10.3384/nejlt.2000-1533.2022.3478}
  {Contextualized embeddings for semantic change detection: Lessons learned}.
\newblock In \emph{Northern European Journal of Language Technology, Volume 8},
  Copenhagen, Denmark. Northern European Association of Language Technology.

\bibitem[{Lesk(1986)}]{lesk1986automatic}
Michael Lesk. 1986.
\newblock Automatic sense disambiguation using machine readable dictionaries:
  how to tell a pine cone from an ice cream cone.
\newblock In \emph{Proceedings of the 5th annual international conference on
  Systems documentation}, pages 24--26.

\bibitem[{{Levenshtein}(1966)}]{levenshtein}
V.~I. {Levenshtein}. 1966.
\newblock {Binary Codes Capable of Correcting Deletions, Insertions and
  Reversals}.
\newblock \emph{Soviet Physics Doklady}, 10:707.

\bibitem[{Mickus et~al.(2022)Mickus, Van~Deemter, Constant, and
  Paperno}]{mickus-etal-2022-semeval}
Timothee Mickus, Kees Van~Deemter, Mathieu Constant, and Denis Paperno. 2022.
\newblock \href {https://doi.org/10.18653/v1/2022.semeval-1.1} {{S}emeval-2022
  task 1: {CODWOE} {--} comparing dictionaries and word embeddings}.
\newblock In \emph{Proceedings of the 16th International Workshop on Semantic
  Evaluation (SemEval-2022)}, pages 1--14, Seattle, United States. Association
  for Computational Linguistics.

\bibitem[{Muennighoff et~al.(2023)Muennighoff, Wang, Sutawika, Roberts,
  Biderman, Le~Scao, Bari, Shen, Yong, Schoelkopf, Tang, Radev, Aji, Almubarak,
  Albanie, Alyafeai, Webson, Raff, and
  Raffel}]{muennighoff-etal-2023-crosslingual}
Niklas Muennighoff, Thomas Wang, Lintang Sutawika, Adam Roberts, Stella
  Biderman, Teven Le~Scao, M~Saiful Bari, Sheng Shen, Zheng~Xin Yong, Hailey
  Schoelkopf, Xiangru Tang, Dragomir Radev, Alham~Fikri Aji, Khalid Almubarak,
  Samuel Albanie, Zaid Alyafeai, Albert Webson, Edward Raff, and Colin Raffel.
  2023.
\newblock \href {https://doi.org/10.18653/v1/2023.acl-long.891} {Crosslingual
  generalization through multitask finetuning}.
\newblock In \emph{Proceedings of the 61st Annual Meeting of the Association
  for Computational Linguistics (Volume 1: Long Papers)}, pages 15991--16111,
  Toronto, Canada. Association for Computational Linguistics.

\bibitem[{Periti and Tahmasebi(2024)}]{periti2024systematic}
Francesco Periti and Nina Tahmasebi. 2024.
\newblock \href {https://doi.org/10.18653/v1/2024.naacl-long.240} {A systematic
  comparison of contextualized word embeddings for lexical semantic change}.
\newblock In \emph{Proceedings of the 2024 Conference of the North American
  Chapter of the Association for Computational Linguistics: Human Language
  Technologies (Volume 1: Long Papers)}, pages 4262--4282, Mexico City, Mexico.
  Association for Computational Linguistics.

\bibitem[{Schlechtweg et~al.(2020)Schlechtweg, McGillivray, Hengchen,
  Dubossarsky, and Tahmasebi}]{schlechtweg-etal-2020-semeval}
Dominik Schlechtweg, Barbara McGillivray, Simon Hengchen, Haim Dubossarsky, and
  Nina Tahmasebi. 2020.
\newblock \href {https://doi.org/10.18653/v1/2020.semeval-1.1}
  {{S}em{E}val-2020 task 1: Unsupervised lexical semantic change detection}.
\newblock In \emph{Proceedings of the Fourteenth Workshop on Semantic
  Evaluation}, pages 1--23, Barcelona (online). International Committee for
  Computational Linguistics.

\bibitem[{Schlechtweg et~al.(2021)Schlechtweg, Tahmasebi, Hengchen,
  Dubossarsky, and McGillivray}]{schlechtweg-etal-2021-dwug}
Dominik Schlechtweg, Nina Tahmasebi, Simon Hengchen, Haim Dubossarsky, and
  Barbara McGillivray. 2021.
\newblock \href {https://doi.org/10.18653/v1/2021.emnlp-main.567} {{DWUG}: A
  large resource of diachronic word usage graphs in four languages}.
\newblock In \emph{Proceedings of the 2021 Conference on Empirical Methods in
  Natural Language Processing}, pages 7079--7091, Online and Punta Cana,
  Dominican Republic. Association for Computational Linguistics.

\bibitem[{Tang et~al.(2023)Tang, Zhou, Aida, Sen, and
  Bollegala}]{tang-etal-2023-word}
Xiaohang Tang, Yi~Zhou, Taichi Aida, Procheta Sen, and Danushka Bollegala.
  2023.
\newblock \href {https://doi.org/10.18653/v1/2023.findings-emnlp.231} {Can word
  sense distribution detect semantic changes of words?}
\newblock In \emph{Findings of the Association for Computational Linguistics:
  EMNLP 2023}, pages 3575--3590, Singapore. Association for Computational
  Linguistics.

\bibitem[{Vijayakumar et~al.(2018)Vijayakumar, Cogswell, Selvaraju, Sun, Lee,
  Crandall, and Batra}]{divbeam}
Ashwin~K. Vijayakumar, Michael Cogswell, Ramprasaath~R. Selvaraju, Qing Sun,
  Stefan Lee, David~J. Crandall, and Dhruv Batra. 2018.
\newblock Diverse beam search: Decoding diverse solutions from neural sequence
  models.
\newblock \emph{AAAI}.

\end{thebibliography}

\newpage

\appendix

\section{Definition generator models}
\label{sec:model_desc}

Contextual definitions in this work are created with fine-tuned large language models, using the method proposed by \citet{giulianelli-etal-2023-interpretable} and further developed in \citet{kutuzov-etal-2024-enriching-word}: an encoder-decoder language model is fine-tuned on a dataset of target word usages and the corresponding definitions. Then, definitions are conditionally generated for every example in the test set. \citet{giulianelli-etal-2023-interpretable} used \texttt{Flan-T5} \cite{chung2022scaling} as the underlying language model. However, it was trained predominantly on English and lacks the capability to properly encode or generate texts in languages with significantly different writing systems (especially true for Russian, but Norwegian characters `å', `ø' and `æ' are also not processed by the \texttt{Flan-T5}  tokenizer).  Because of that, \citet{kutuzov-etal-2024-enriching-word} and this work are using \texttt{mT0-xl}~\cite{muennighoff-etal-2023-crosslingual} which is essentially a multilingual version of \texttt{Flan-T5}. 
Fine-tuning was done in a standard text-to-text setup, for every language (English, Norwegian, Russian) separately, so that in the end we had three language-specific models: \url{https://huggingface.co/ltg/mt0-definition-en-xl}, \url{https://huggingface.co/ltg/mt0-definition-no-xl} and \url{https://huggingface.co/ltg/mt0-definition-ru-xl}.

\section{Examples of merged definitions}
\label{sec:defex}

Tables~\ref{tab:eng_examples}, \ref{tab:nor_examples} and \ref{tab:rus_examples} show examples of most frequent definitions-as-senses for some of the target words in our English, Norwegian and Russian benchmarks. All the definitions are generated with the best strategy for the specific dataset (see Table~\ref{tab:merge}).

\begin{table*}
\begin{tabular}{p{13mm}p{67mm}p{67mm}}
\toprule
& \multicolumn{1}{c}{\textbf{Period 1} (1810-1860)} & \multicolumn{1}{c}{\textbf{Period 2} (1960-2010)} \\
\midrule
\multirow[t]{3}{=}{\textit{circle} JS=0.07} & \textit{To move in a circular course} (100\%) & \textit{To move in a circular course} (99\%) \\
\cmidrule{2-3}
& & \textit{To move around something especially so as to make it appear to move around} ($<$1\%)\\
\cmidrule{2-3}
& & \textit{To move around an axis or centre especially as if following a regular path around an axis} ($<$1\%)\\
\midrule
\multirow[t]{3}{=}{\textit{risk} JS=0.44} & \textit{The probability of a negative outcome to a decision or event} (59\%) & \textit{The probability of a negative outcome to a decision or event} (63\%) \\
\cmidrule{2-3}
 & \textit{The probability of a negative outcome to a decision or event  the chance of a negative outcome to a decision or event} (8\%) & \textit{The probability of a negative outcome to a decision or event  the chance of a negative outcome to a decision or event} (3\%) \\
\cmidrule{2-3}
& \textit{A venture undertaken without regard to possible loss or injury especially if significant} (3\%) & \textit{A venture undertaken without regard to possible loss or injury especially if significant} (3\%) \\
\midrule
\multirow[t]{3}{=}{\textit{ball} JS=0.83} & \textit{A spherical object especially one that is round in shape} (82\%) & \textit{The object hit in a game} (80\%) \\
\cmidrule{2-3}
 & \textit{A party} (6\%) & \textit{The object used in various sports especially in soccer tennis basketball etc} ($<$1\%) \\
\cmidrule{2-3}
& \textit{A wedding} ($<$1\%) & \textit{The object used in various sports especially in soccer basketball and other games which is thrown or kicked} ($<$1\%)\\
\bottomrule
\end{tabular}
\caption{The three most frequent definitions per period for three English words: \textit{circle} (low predicted change rate), \textit{risk} (medium predicted change rate), and \textit{ball} (high predicted change rate). Parentheses indicate the relative frequency of each definition among all samples of the period.}
\label{tab:eng_examples}
\end{table*}

\begin{table*}
\begin{tabular}{p{17mm}p{65mm}p{65mm}}
\toprule
& \multicolumn{1}{c}{\textbf{Period 1} (1980-1990)} & \multicolumn{1}{c}{\textbf{Period 2} (2012-2019)} \\
\midrule
\multirow[t]{3}{=}{\textit{oppvarming} `heating, warm-up' JS=0.19} & \textit{det å varme opp} `the action of heating' (98\%) & \textit{det å varme opp} `the action of heating'(91\%) \\
\cmidrule{2-3}
& \textit{i regnskap} `in accounting' ($<$1\%) & \textit{det å varme opp jordoverflaten} `the action of warming the Earth surface' (1\%)\\
\cmidrule{2-3}
& \textit{i statistikk} `in statistics' ($<$1\%) & \textit{i fotball} `in football' ($<$1\%)\\
\midrule
\multirow[t]{3}{=}{\textit{bank} `bank' JS=0.64} & \textit{institusjon som tar imot innskudd av penger og gir lån} (13\%) &  \textit{institusjon som tar imot innskudd av penger og gir lån} (14\%) \\
& \multicolumn{2}{c}{`institution that accepts money deposits and gives loans'} \\
\cmidrule{2-3}
& \textit{institusjon som tar imot innskudd og utfører pengetransaksjonstjenester} (6\%) & \textit{institusjon som tar imot innskudd og utfører pengetransaksjonstjenester} (6\%) \\
& \multicolumn{2}{c}{`institution that accepts deposits and provides financial transaction services'} \\
\cmidrule{2-3}
& \textit{institusjon som tar imot innskudd av penger og driver pengetransaksjonsvirksomhet} (5\%) & \textit{institusjon som tar imot innskudd av penger og driver pengetransaksjonsvirksomhet} (4\%) \\
& \multicolumn{2}{c}{`institution that accepts deposits of money and conducts financial transaction business'} \\
\midrule
\multirow[t]{3}{=}{\textit{kode} `code' JS=0.81}  & \textit{i i sammensetninger} `i in compounds' (4\%) & \textit{i i bestemt form} `i in the definite form' (4\%) \\
\cmidrule{2-3}
& \textit{i i bestemt form} `i in the definite form' (3\%) & \textit{mønster oppskrift på hvordan noe skal lykkes} `pattern, recipe for how something succeeds` (1\%) \\
\cmidrule{2-3}
& \textit{i statistikk} `in statistics' (3\%) & \textit{i overført betydning mønster mønstergyldighet} `in a figurative sense pattern, pattern validity' (1\%)\\
\bottomrule
\end{tabular}
\caption{The three most frequent definitions per period for three words of the Norwegian-2 dataset: \textit{oppvarming} (low predicted change rate), \textit{bank} (medium predicted change rate), and \textit{kode} (high predicted change rate). Parentheses indicate the relative frequency of each definition among all samples of the period.}
\label{tab:nor_examples}
\end{table*}

\begin{table*}
\begin{tabular}{p{15mm}p{67mm}p{67mm}}
\toprule
& \multicolumn{1}{c}{\textbf{Period 1} (before 1917)} & \multicolumn{1}{c}{\textbf{Period 2} (after 1991)} \\
\midrule
\multirow[t]{3}{=}{\foreignlanguage{russian}{\textit{цензура}} \textit{`censorship'} JS=0.09} & \begin{otherlanguage*}{russian}\textit{система государственного надзоря за печатью и средствами массовой информации}\end{otherlanguage*} (99\%) & \begin{otherlanguage*}{russian}\textit{система государственного надзоря за печатью и средствами массовой информации}\end{otherlanguage*} (99\%) \\
& \multicolumn{2}{c}{\textit{system of state control over printing and mass media}} \\
\cmidrule{2-3}
  & \begin{otherlanguage*}{russian}\textit{истор. государственный орган, осуществляющий цензуру}\end{otherlanguage*} \textit{historically, a state body conducting censorship} ($<$1\%) & \begin{otherlanguage*}{russian}\textit{государственная система государственного надзоря за печатью и средствами массовой информации}\end{otherlanguage*} \textit{a state system of controlling printing and mass media} ($<$1\%)\\
\cmidrule{2-3}
& \begin{otherlanguage*}{russian}контроль, надзор за печатью и средствами массовой информации\end{otherlanguage*} \textit{control and monitoring of printing and mass media} ($<$1\%)  & \begin{otherlanguage*}{russian}\textit{сокр. от цензурная служба,  государственная организация, осуществляющая цензуру}\end{otherlanguage*} \textit{abbrev. censoring body, state organ conducting censorship} ($<$1\%)\\
\midrule
\multirow[t]{3}{=}{\foreignlanguage{russian}{\textit{огонь}} \textit{`fire'} JS=0.66} & \begin{otherlanguage*}{russian}\textit{источник огня, источник света
}\end{otherlanguage*} \textit{source of fire or light} (7\%) & \begin{otherlanguage*}{russian}\textit{действие по значению глагола стрелять}\end{otherlanguage*} \textit{nominal form of the verb `to fire'} (3\%) \\
\cmidrule{2-3}
 & \begin{otherlanguage*}{russian}\textit{источник света, источник тепла, дыма и т.п.}\end{otherlanguage*} \textit{source of light, warmth, smoke etc} (2\%) & \begin{otherlanguage*}{russian}\textit{источник света, источник освещения}\end{otherlanguage*} \textit{source of light, of illumination} (2.5\%) \\
\cmidrule{2-3}
& \begin{otherlanguage*}{russian}\textit{перен. страсть, пыл}\end{otherlanguage*} \textit{metaphoric. passion or rage} (2\%) & \begin{otherlanguage*}{russian}\textit{воен. стрельба из огнестрельного оружия}\end{otherlanguage*} \textit{metaphoric. gunfire} (1\%) \\
\midrule
\multirow[t]{3}{=}{\foreignlanguage{russian}{\textbf{линейка}} \textit{`line, ruler'} JS=0.80} & \begin{otherlanguage*}{russian}\textit{измерительный инструмент в виде прямой пластинки с нанесёнными на неё делениями для измерения длины и расстояния}\end{otherlanguage*} \textit{a measuring tool looking like a straight plane with marks to measure length and distance} (2.6\%) & \begin{otherlanguage*}{russian}\textit{перен. совокупность однородных предметов, изделий, продуктов и т. п.}\end{otherlanguage*} \textit{metaphoric. a batch of similar items, goods, products} (3.5\%) \\
\cmidrule{2-3}
 & \begin{otherlanguage*}{russian}\textit{устар. длинные узкие сани}\end{otherlanguage*} \textit{archaic. long narrow sledges} (1\%) & \begin{otherlanguage*}{russian}\textit{измерительный инструмент в виде прямой пластинки с нанесёнными на неё делениями для определения длины линии}\end{otherlanguage*} \textit{a measuring tool looking like a straight plane with marks to measure the length of a line} (1.4\%) \\
\cmidrule{2-3}
& \begin{otherlanguage*}{russian}\textit{измерительный инструмент в виде прямой линии с нанесёнными на неё делениями}\end{otherlanguage*} \textit{a measuring tool looking like a straight plane with marks} (1\%) & \begin{otherlanguage*}{russian}\textit{измерительный инструмент в виде прямой пластинки с нанесёнными на неё делениями для измерения длины и ширины}\end{otherlanguage*} \textit{a measuring tool looking like a straight plane with marks to measure length and width} (1\%)\\
\bottomrule
\end{tabular}
\caption{The three most frequent definitions per period for three words from the Russian-1 dataset: `\foreignlanguage{russian}{\textit{цензура}}' (low predicted change rate), `\foreignlanguage{russian}{\textit{огонь}}' (medium predicted change rate), and  `\foreignlanguage{russian}{\textit{линейка}}' (high predicted change rate). Parentheses indicate the relative frequency of each definition among all samples of the period.}
\label{tab:rus_examples}
\end{table*}

\end{document}